\documentclass[10pt,letterpaper]{article}
\usepackage{algorithm}
\usepackage{algpseudocode}
\usepackage{cogsci}
\usepackage{graphicx}
\usepackage{placeins}

\cogscifinalcopy 

\usepackage{cmap}
\usepackage[T1]{fontenc}
\usepackage[american]{babel}
\usepackage{csquotes}
\usepackage{newtxtext,newtxmath}  

\usepackage{natbib}

\usepackage{float} 

\usepackage[hidelinks]{hyperref}

\title{Path Integration and Object-Location Binding Emerge in\\an Action-Conditioned Predictive Sequence Network}

\author[*1, 2]{\mbox{Linda Ariel Ventura}}
\author[*1]{\mbox{Victoria Bosch}}
\author[1]{\mbox{Tim C Kietzmann}}
\author[1]{\mbox{Sushrut Thorat}}
\affil[1]{Institute of Cognitive Science, Osnabrück University, Germany}
\affil[2]{Sorbonne University, France}
\affil[*]{Shared first authorship}

\begin{document}

\maketitle
\vspace{-1em}

\begin{abstract}
Adaptive cognition requires structured internal models of objects and their relations. Predictive neural networks are often proposed to learn such world models, but how these are instantiated and how they support prediction remain unclear. We investigate this in a minimal in-silico setting. A recurrent neural network samples tokens sequentially from 2D continuous token scenes and is trained to predict the upcoming token from the current input and a saccade-like displacement. On novel scenes, prediction accuracy improves across the sequence, indicating in-context learning. Decoding analyses reveal path integration and dynamic binding of token identity to position. Interventional analyses show that new bindings can be learned late in sequence and that out-of-distribution bindings can be learned as well. Together, these findings show how structured representations relying on flexible binding emerge to support prediction, offering a mechanistic account of sequential world modeling relevant to cognitive science.\footnote{The code to reproduce these results can be found at:\\ \url{https://github.com/KietzmannLab/minimal_world_model_interp}}

\end{abstract}

\section{Introduction}

\begin{figure*}[t]

    \centering

    \includegraphics[width=0.97\textwidth]{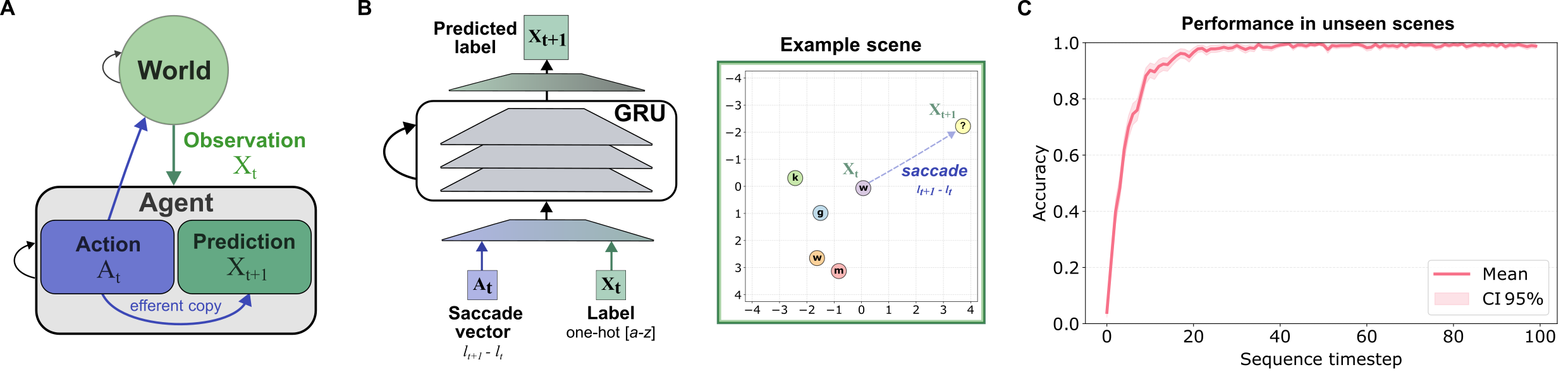}
    \caption{\textbf{A minimal action-conditioned predictive sequence network.} \textbf{A.} Conceptual figure depicting action-conditioned prediction. \textbf{B.} Our instantiation of a prediction network, and an example of a scene with $6$ tokens placed in a $2$D continuous space. The network receives the saccade displacement vector and the label of the current token, and is tasked to output the label of the subsequent token in the sequence. \textbf{C.} Average network prediction performance in unseen worlds (N=$500$) per sequence timestep. The network quickly learns the structure of novel worlds, using its dynamics without any parameter changes (in-context learning). Error envelope depicts $95\%$ CIs of the means.
}
    \label{fig1}
\end{figure*}

Understanding and operating in the natural world requires representing objects and their relations, and updating those representations as new information is acquired. How could biological agents acquire such structured models of the world? A notable proposal in cognitive science and machine learning is that such structured internal world models can emerge from learning to predict future sensory inputs, notably when prediction is conditioned on the agent’s own actions (i.e., efference copies; see Figure~\ref{fig1}A)~\citep{elman1990finding, o2001sensorimotor, ha2018world, petroni2019language, brown2020language, lecun2022path}. For example, cognitive science accounts of sensorimotor contingencies and affordance-based cognition emphasize that perception is structured by the expected sensory consequences of possible actions~\citep{o2001sensorimotor, eperon2026action}. While both biological agents and predictive neural networks appear capable of acquiring structured models, it remains unclear how these ``world models'' are implemented internally: what mechanisms encode the sensed parts of the world and their relationships, and retrieve the relevant parts for prediction?

One recent example of a system performing action-conditioned prediction is the Glimpse Prediction Network (GPN), where predicting the content of the next fixation in a sequence of eye movements, conditioned on a saccadic efference copy, drives integration across time and yields unified scene representations aligned with human neural responses to natural scenes~\citep{thorat2025predicting}. To probe the mechanisms supporting such model-based sequence prediction, we study a minimal setting inspired by GPN. Our goal is to retain the core ingredients of such action-conditioned prediction, while stripping away domain-specific complexity (e.g., visual co-occurrence and semantics of scene parts). This allows for increased interpretability and ease of probing the network's internal mechanisms. Here, we consider scenes as sets of tokens in a two-dimensional continuous space, and construct sequences of displacements between tokens (saccades). At each step, a recurrent neural network (RNN) predicts the label of the next token given the current token and the provided displacement.

Training to predict the next token in sequences over many scenes induces an in-context learning ability to encode tokens and their relative positions in novel scenes, without any weight updates. Because the latent structure of the scenes is fully known, this setting allows precise interrogation of the network's internal representations of the scene components and their relations. To identify mechanisms and decompose computational components underlying this behavior, we formulate a hypothesis space using a symbolic algorithm for saccade-conditioned token encoding and retrieval. We show that the hypothesized algorithmic components exist in the network: path integration of saccades and binding of tokens to absolute positions. Interventional analyses demonstrate that the network can memorize new label-position bindings in-context while retaining previously stored bindings, and the binding operation extends to out-of-distribution token-position pairs. Together, these results demonstrate how an action-conditioned prediction objective can give rise to mechanisms that implement a structured model of the observed world. More broadly, this work demonstrates how mechanistic interpretability analyses of a minimal network can reveal algorithmic components that might underlie world modeling in more complex predictive systems.

\FloatBarrier
\section{Methods}

\subsection{Minimal scene construction}

As seen in Figure ~\ref{fig1}B, each minimal scene consists of $4-6$ tokens sampled from the $26$ possible letters of the English alphabet (a-z); letters can occur multiple times. These tokens are placed on a continuous 2D space with x/y coordinate bounds of $[-4,4]$ and a minimal distance of $0.25$ between tokens. As a result, the space of possible label-position combinations is massive, allowing for on-the-fly data generation during training without substantial overlap between examples. Saccade sequences are sampled from these scenes: at any given timestep, a random displacement to one of the other tokens is initiated. The first timestep always corresponds to the token at the center of the scene.

\subsection{Network architecture \& training}

As seen in Figure~\ref{fig1}B, the network that we tasked with this simplified next-token prediction is a $3$-layer Gated-Recurrent Unit (GRU) RNN~\citep{cho2014learning,paszke2019pytorch}, with a hidden state size of $512$. The $26$D one-hot token label input and 2D saccade (displacement between two tokens) input are linearly projected to a $512$D layer, which is fed as input to the GRU. The output of the GRU is projected to a $512$D ReLU layer, from which a $26$D linear readout predicts the upcoming token label at the next timestep. We train the network until convergence (for $40960$ batches of $200$ scenes each; one sequence per scene; sequence length of $100$ timesteps), using cross-entropy loss. As any token can be present at any position in the scenes, we expect the network to rely on its internal dynamics to learn about the token distributions in the scene being sensed through the input sequence.

\subsubsection{Robustness}
Results reported in this paper were qualitatively reproduced with two more different random seed initializations, as well as in a 3-layer parameter-matched LSTM. All statistics reported in this manuscript are acquired using two-sided t-tests. 

\section{Results}
\subsection{In-context learning of token arrangement in scenes}

After training the recurrent neural network to predict the next token in sequences over many scenes, we probe its generalization capacity by testing the trained frozen network on sequences in newly-generated scenes. The network predicts the upcoming token with increasing accuracy as the sequence proceeds (Figure~\ref{fig1}C; N = $500$ scenes). The network thus demonstrates capacity for in-context learning of the token arrangement in scenes~\citep{olsson2022context}, and reaches peak prediction accuracy within $35$ timesteps. 

Next, we ask how the network is capable of learning a ``world model'', i.e., the tokens and their relative positions in scenes, and what format and mechanisms it uses to store this knowledge for the retrieval of the next token without changing its weights. 

\begin{algorithm}
\caption{Hypothesized Symbolic Algorithm}
\label{alg:symbolic}
\begin{algorithmic}[1]
\State Initialize dictionary $D \leftarrow \emptyset$ \& position $p \leftarrow (0,0)$

\For{$t = 1$ to $T$}
    \State Observe token label $X_t$ and intended saccade $A_t$
    \State Store $D[p] \leftarrow X_t$
    \State Update position: $p \leftarrow p + A_t$
    \If{$p$ is a key in $D$}
        \State Output stored label $D[p]$
    \Else
        \State Output a random label from alphabet $\mathcal{A}$
    \EndIf
\EndFor
\end{algorithmic}
\end{algorithm}

\section{Signatures of path integration and token label-position binding}

As a starting point, we hypothesize a symbolic algorithm that the network could implement (Algorithm~\ref{alg:symbolic}). The network must memorize observed tokens and process their relative positions. A general memory-efficient solution is a dictionary-like format that binds positions to labels, i.e. \textit{\{position: label\}}. Compared with a transition cache of local tuples, \textit{(label1, saccade, label2)}, this format requires $O(N)$ rather than $O(N^2)$ memory, where $N$ is the number of tokens. Crucially, a bound label-position format allows post-hoc inference of unseen token relations, as required when any token can be queried by a saccade. Consistent with this, the network zero-shot infers the true next token for saccades withheld for the first $100$ timesteps ($99.2\%$ accuracy; N = $500$ scenes), suggesting that it encodes absolute token positions. This favors position-based binding over a transition cache, which would not support such inference.

\begin{figure*}[t]
    \includegraphics[width=\textwidth]{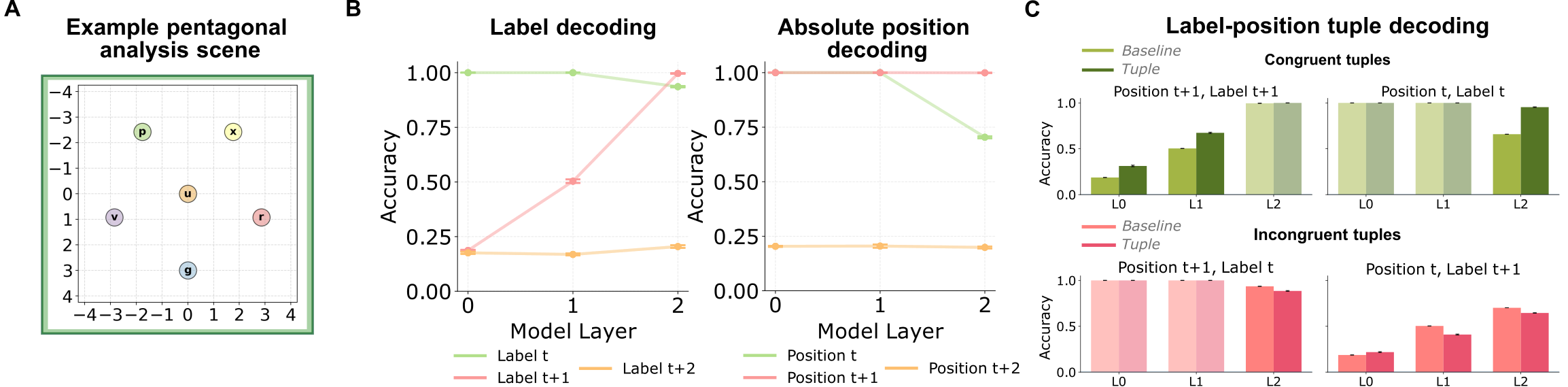}
    \caption{\textbf{Assessing algorithmic components in the network's representations.} \textbf{A.} Example pentagonal scene used for the decoding analyses. We generate $500$ test scenes with this arrangement. \textbf{B.} Decoding of token label and absolute token position across network layers, averaged across timesteps $35-100$. High absolute position decoding suggests the network learned path-integration. \textbf{C.} Decoding of label-position tuples across network layers, averaged across timesteps $35-100$, compared to their corresponding baselines, i.e., the expected decoding accuracy based on the decoding of the components (product of their decoding accuracies). The top panels show results for congruent label-position pairs, and the bottom panels depict results for incongruent label-position pairs. We highlight the results for which the baseline accuracy is below $1$, as in the cases of perfect component decoding we cannot meaningfully measure accuracy deviations that would indicate binding. The congruent tuples, but not the incongruent tuples, can be decoded better than their corresponding baselines, suggesting that the positions of tokens were bound to their labels resulting in a distinct representation compared to the parts. Error bars depict $95\%$ CIs of the means.}
    \label{fig2}
\end{figure*}

To construct a bound label-position representation, we hypothesize two capabilities are necessary: 1) the network needs to path integrate sequentially to infer the absolute position at a given timestep, and 2) bind the currently-seen token to the inferred absolute position and store it. A final component for the prediction is retrieval: at a timestep, given the saccade and performing path integration, the network has to be able to retrieve the label of the next token at the inferred position, and output the predicted token. To search for the components of such an algorithm, we set up a controlled scene, sampling 6 tokens and arranging them as a pentagon (radius of the circumscribing circle = $3$) and its center (see Figure~\ref{fig2}A; 500 scenes generated for testing the network). We commence by asking whether we can decode the token labels and positions at different timesteps from the layer activations of the network (across sequence timesteps $35-100$), using Support Vector Machine (SVM) classifiers with $5$-fold cross-validation~\citep{cortes1995support,pedregosa2011scikit}. We decode token label ($26$-way classification) and absolute position ($6$-way classification) at the current timestep ($t$), as well as the two subsequent timesteps ($t+1$ and $t+2$). Here, $t+2$ acts as a baseline, as neither token label nor position for this timestep can be known from the history, currently available, or predicted information. Indeed, we observe chance performance ($20\%$) throughout the layers for both label and position decoding at timestep $t+2$.

Given the hypothesized processes related to binding and retrieval, we expect the current and predicted token identities to be represented in the network layers. Indeed, as seen in Figure~\ref{fig2}B (left), we observe high label decoding for the current token in all layers, whereas the accuracy for the next token increases with layers throughout the network. This result is intuitive, because the current token is provided as input, whereas the next token has to be inferred and serves as the output of the model.

The path integration hypothesis predicts both the current and predicted token positions to be represented in the network layers. We observe perfect absolute position decoding for both timesteps in the first layer, and a small decrease in accuracy with layer depth, more so for the current position than the predicted position (Figure~\ref{fig2}B; right). High linear separability of both the current and subsequent absolute positions signals path integration, as the network only receives relative token positions (saccades).

\begin{figure*}[t]
    \centering
    \includegraphics[width=\textwidth]{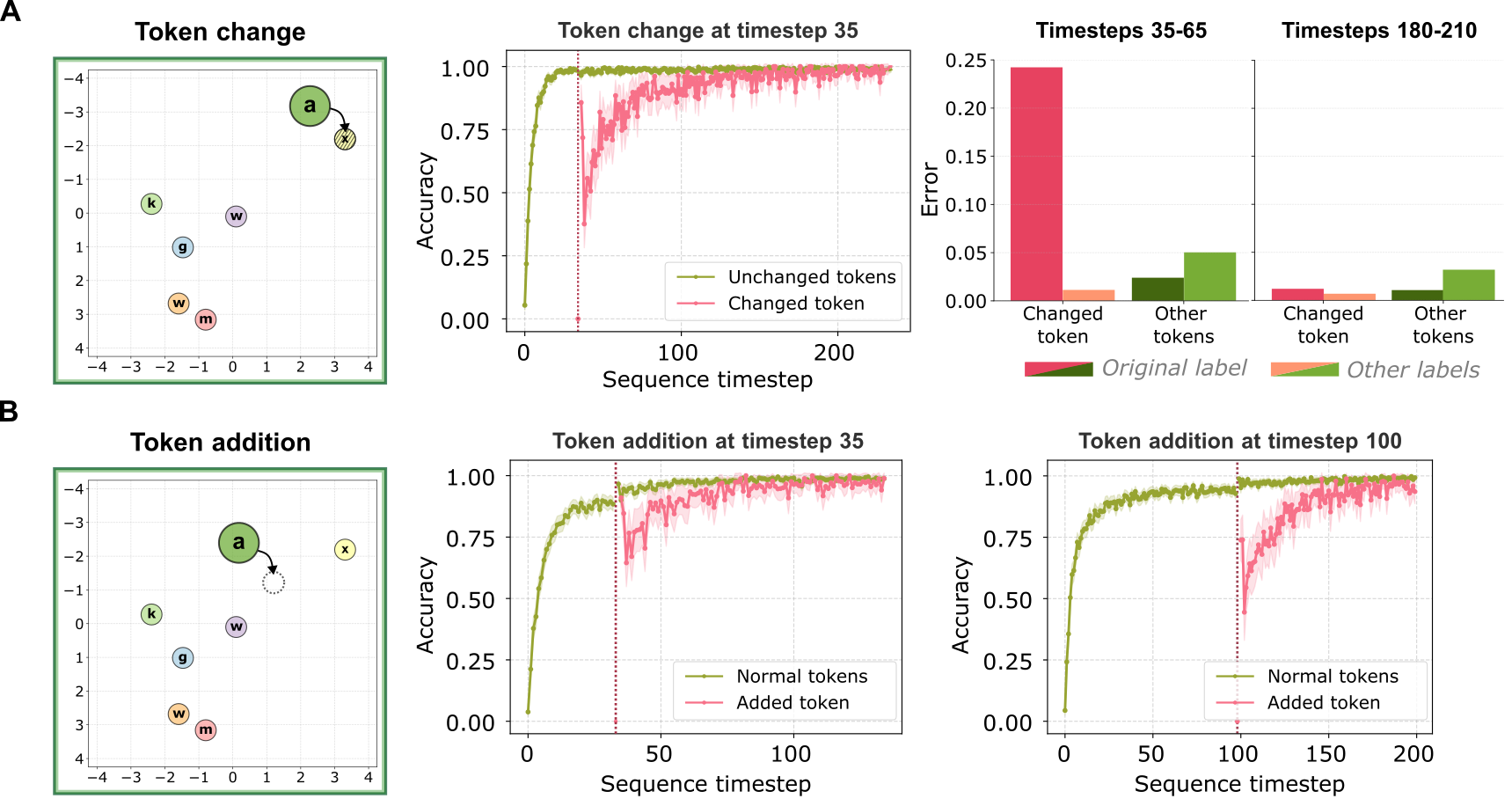}
    \caption{\textbf{Assessing the stability and plasticity of in-context memory.} \textbf{A.} Left: Illustration of the token change procedure. Middle: Performance under intervention for tokens at the changed and unchanged locations (N$=500$ scenes). Right: Error contribution of original replaced labels and other labels at other locations. When a token at a particular location is changed, the network gradually switches to predicting the new token at that location, suggesting short-timescale stability of its memory. \textbf{B.} Left: Illustration of the token addition procedure. Middle: network performance after token addition at timestep $35$ (N$=500$ scenes). Right: network performance after token addition at timestep $100$ (N$=500$ scenes). When a new token is added at a new location, the network requires multiple repetitions of exposure to encode it in memory. It can do so even after prior heavy exposure to the rest of the scene, suggesting a consistently plastic memory. Error envelopes depict $95\%$ CIs of the means.}
    \label{fig3}
\end{figure*}

Having confirmed that the components (token label and position) are present in the network representations, we move on to search for evidence of them being bound together. We test whether the network encodes bound representations of label and absolute position, i.e., whether congruent label-position tuples are more decodable ($26\times 6$ = $156$-way classification) than would be expected from a joint-decoding baseline, which is the product of the decoding accuracies for the token label and the position separately. Note that if the components are perfectly decodable, i.e, baseline accuracy = $1$, we cannot meaningfully determine whether these components are bound, as the expected tuple decoding will also be $1$. As shown in Figure~\ref{fig2}C (top panels), decoding accuracy for congruent tuples exceeds their baselines: for the tuple at the upcoming position (timestep $t+1$), decoding accuracy exceeds the baseline at the first two layers (both $p<0.001$). For the tuple at the current position (timestep $t$), decoding accuracy is above baseline in layer $2$ ($p<0.001$).

To rule out the possibility that \textit{any} combination of label and position is jointly decodable above baseline, we run a mismatch control. Specifically, we consider cross-timestep tuples by combining the position (or label) at the current timestep with the label (or position) from the subsequent timestep, decoding these incongruent combinations, and comparing them with their corresponding product baselines. Unlike the congruent tuples, these incongruent tuples show little to no elevation, or a reduction compared to their baselines (Figure~\ref{fig2}C; all layers differ significantly from baseline, $p<0.001$). Thus, the joint-decoding boost is selective for the expected (congruent) label-position pairing, consistent with a specific position-label bound representation rather than a generic mixture of decodable components. This provides evidence towards the hypothesized mechanism of binding of label and position in the network. 

We reproduce these results in a parameter-matched LSTM where joint decoding of the label-position tuple is better than baseline for congruent tuples (position $t+1$: $p<0.001$ for all layers; position $t$: $p<0.001$ for L2) and baseline decoding improves over incongruent tuples (position $t+1$: $p<0.001$ for L2; position $t$: $p<0.001$ for L0, L1). This indicates that it is plausible that other architectures may solve the task using similar algorithmic components.

\section{The stability, plasticity, and generalizability of in-context scene memory}

To further characterize the network’s in-context learning ability and our finding that its memory contains bound representations, we use interventional analyses to ask at what stage in the sequence these representations can be introduced or modified, and whether out-of-distribution label-position bindings can be learned.

First, we ask whether we can replace a token after it has been memorized at a position. As we observe convergence of prediction accuracy around timestep $35$, we chose to replace one of the scene’s tokens at that time point and continue the sequence accordingly (see Figure~\ref{fig3}A, left panel; N = $500$ scenes with $6$ tokens each). Measuring the prediction accuracy of the network for the tokens at both the unchanged positions as well as the (new) token at the changed position, we observe that the performance at other positions does not change after the intervention, indicating no observable disruption of the memory representations (Figure~\ref{fig3}A, middle panel). For the changed position, we do observe a gradual increase in prediction performance of the new token across sequence steps. To better understand the memory encoding and process underlying these observations, we further investigate the causes of error on the changed position: is the original label-position tuple \textit{overwritten}, or does it keep competing with the newly-introduced tuple? Separating the cause of error by token label, we observe that initially (between timesteps $35-65$), the largest share of the errors made on the position of intervention are due to the network erroneously predicting the original label (see Figure~\ref{fig3}A, right panel). However, after $180$ time steps, this error type decreases significantly in occurrence ($p<0.001$), revealing that although challenging, increased exposure of the new token at this position does overwrite the original memory of the label-position tuple. 

Next, we ask whether we can add in new tokens after the network has converged on its performance on a given test scene. We do so by introducing a novel token at a later timestep to $500$ scenes with 5 randomly-picked tokens each (Figure~\ref{fig3}B, left panel), and evaluating performance thereafter (while randomly cycling through the $6$ tokens). After introducing the new token at timestep $35$, the network quickly learns to start predicting the new token at its position, reaching peak accuracy after $\sim50$ steps (Figure~\ref{fig3}B, middle panel). Does the network’s memory become less plastic over time? We observe that, when introducing the new token at timestep $100$, the network is still able to learn the new token and predict it with high accuracy after $\sim50$ steps as well (Figure~\ref{fig3}B, right panel), without changing its weights, indicating that in-context memory plasticity does not reduce over time.

\begin{figure}[t]
    \centering
    \includegraphics[width=0.9\columnwidth]{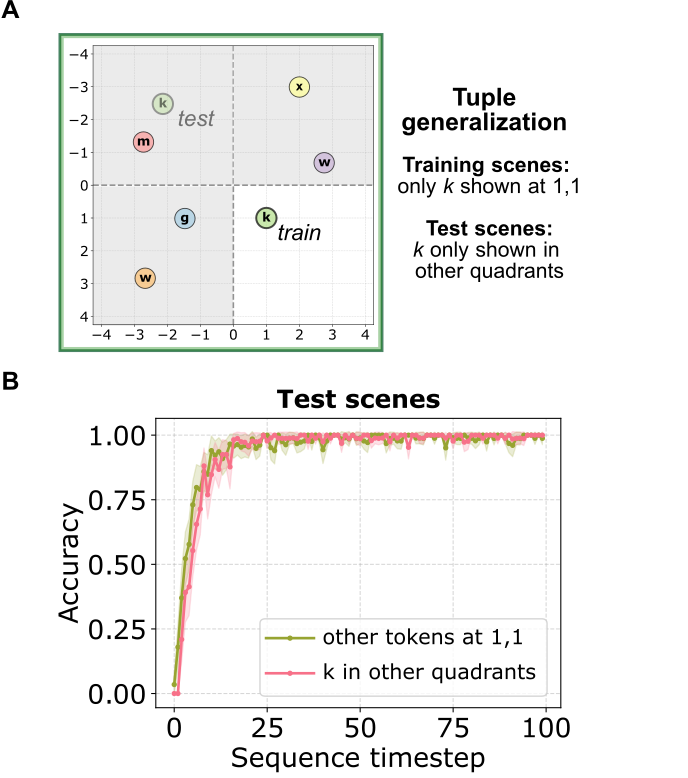}
    \caption{\textbf{Assessing the generalizability of token identity-location binding.} \textbf{A.} Illustration of tuple generalization analysis: during training, the token \textit{k} is only shown at position $(1,1)$ while other tokens can be placed anywhere on the grid. \textbf{B.} Network performance on test scenes with a non-\textit{k} token placed at the control location $(1,1)$, and \textit{k} placed in the other three quadrants (N$=500$ scenes). These results suggest that the network can flexibly bind tokens to locations, overriding the label-location co-occurrence learned during training. Error envelope depicts $95\%$ CIs of the means.}
    \label{fig4}
\end{figure}

Finally, as a test of the generalizability of in-context memory to unseen token arrangements, we ask if the network can learn to associate a token with positions where it was never seen during training. Specifically, we set up a control setting: during network training the label \textit{k} is only ever shown in the lower-right quadrant at the control position $(1,1)$, while no other label is shown at that position or in a $0.25$ radius area around it (see Figure~\ref{fig4}A). After training, we construct test scenes of 6 tokens, in which a \textit{k} token is only shown in the other three quadrants of the scene (the quadrants not containing position $(1,1)$), and one of the other tokens is shown at the control position. We evaluate performance for predicting \textit{k}, and the other token at $(1,1)$. We observe that the network can learn to infer \textit{k} in the three other quadrants, and can infer any other token at the control position (N = $500$ scenes; see Figure~\ref{fig4}B). This suggests that the network can arbitrarily bind known tokens to known positions, even if during training those tokens were never seen in those positions.

Taken together, these interventional analyses illustrate the stability, flexibility, and generalizability of in-context learning in the network. The network is capable of memorizing new tokens at any point in time, and does not show critical phases in its learning process. Yet, it appears to be challenging for the network to overwrite an existing bound representation, signaling stable memory representations. This suggests that the memory representations are more complex than a simple dictionary-like representation. If it were to be dictionary-like, label changes would happen instantaneously, as once the position is indexed, the associated token could be replaced. Finally, the binding operation is not limited to known label-position distributions but can be extended to new, out-of-distribution label-position pairs.

\section{Discussion}

Training a recurrent neural network to predict the next token label from two ingredients (the current token and a saccade displacement) induces a robust in-context learning ability: on newly generated scenes, the frozen network improves next-token prediction as the sequence unfolds, without any weight updates~\citep{brown2020language,olsson2022context}. Because scenes are generated on the fly in continuous space, this improvement cannot be explained by memorizing a finite set of scenes or fixed label-position patterns. Instead, the network must use the sequence itself to infer the tokens present in the current scene and how they are arranged, and use that inferred structure to answer arbitrary next-token queries driven by the saccade input.

Mechanistically, our analyses point to two core ingredients. First, the network represents absolute position despite receiving only relative displacements, consistent with internal integration of saccades across time. Related computational pressures are known to elicit path integration codes in recurrent networks trained on self-motion and localization tasks~\citep{cueva2018emergencegridlikerepresentationstraining,banino2018vector,sorscher2023unified}. Second, beyond representing labels and positions separately, the network shows evidence consistent with the binding of labels and their absolute positions: congruent label-position tuples are selectively more decodable than expected from the components alone, and mismatched controls show that this boost is specific to the correct pairing. Together with correct prediction under saccades not encountered early in the sequence, these findings support the idea that the network forms an internal record akin to ``this label is at this position'' that can be queried for prediction.

Note that \textit{binding} here refers to a retrieval-oriented label-position association rather than classical feature conjunction. In cognitive science, binding often refers to combining separable perceptual attributes into coherent objects~\citep{treisman1980feature,kahneman1992reviewing,scholte2025beyond}. Our task instead demands a role-filler style association: a token identity must be linked to a position so that it can be retrieved when a saccade specifies a target. This requirement speaks to longstanding questions about whether distributed connectionist representations support systematic variable binding~\citep{fodor1988connectionism}, and connects to work on dynamic binding in structured representations, including synchrony-based proposals and distributed binding schemes~\citep{hummel1992dynamic,smolensky1990tensor,plate1995holographic,kanerva2009hyperdimensional,kleyko2022survey,griffiths2025whither}. Future work could assess the \textit{format} of the learned association, for example by testing whether network states are well-approximated by superposed role-filler bindings as in tensor product representations, or by treating alternative vector-symbolic binding operators as competing hypotheses in fits to internal activations~\citep{mccoy2018rnns,plate1995holographic,greff2020binding}.

We note that, a priori, path integration and binding are not the only possible mechanisms that could solve the sequential prediction task. However, our results challenge alternative mechanisms such as a within-episode transition cache that stores local tuples of the form \textit{(label1, saccade, label2)}. Such a cache would be expected to fail when queried with unseen saccades, whereas we observe robust zero-shot inference under withheld displacements. Moreover, the results of the out-of-distribution manipulation are better explained by a compositional binding mechanism, because it directly tests whether binding depends on label-specific training ``support'' (e.g., a label being bindable only at positions encountered during training). The network's success indicates that the association operation can generalize to new positions for a label even when its training exposure was spatially restricted. These results reinforce our hypothesis that the mechanisms underlying the network's capacities rely on structured and relational mechanisms, rather than simpler lookup-based strategies. 

Interventional analyses further demonstrated that the in-context scene memory encoded by the network is both plastic and stable. New label-position pairings can be acquired late in a sequence, yet overwriting an existing pairing is slow and initially dominated by the original label. This overwrite difficulty argues against a simple dictionary-like memory format and suggests the persistence of old associations that interfere with new ones, echoing classic constraints in connectionist accounts of learning and memory~\citep{mccloskey1989catastrophic,mcclelland1995there}.

The mechanisms identified here also connect to proposed hippocampal-entorhinal mechanisms for spatial and relational memory. Path integration and flexible item-location binding are central to accounts of cognitive maps and are explicit in models such as the Tolman-Eichenbaum Machine, where relational structure is learned from sequential experience and generalized to new queries~\citep{whittington2020tolman}. This raises the possibility that related algorithmic components support biological sequence learning. Testing this would require a comparable task in which subjects learn a novel layout through sequential observations and action-related signals, while probing neural activity recordings for integrated position, upcoming position, item identity, and item-position pairings. Such a comparison is nontrivial: efference copies help predict the sensory consequences of action~\citep{wolpert1995internal,wolpert2001motor}, and action and perception are often coupled rather than separable~\citep{friston2011action}. Nevertheless, such experiments could directly test whether analogous path-integration and binding operations support biological sequence modeling.

In sum, this minimal setting enables identification of candidate algorithmic components (path integration and label-position binding) that support action-conditioned sequential prediction. By retaining only the core objective of action-conditioned prediction, this minimal setting serves as a pathway for mechanistic interpretability: it narrows the space of plausible solutions while still producing nontrivial internal structure. This provides a concrete starting point for asking how such components are implemented in recurrent dynamics (or through attention in transformers), and whether analogous mechanisms arise in more complex active-vision models and in biological systems that tie together perception and action through prediction~\citep{o2001sensorimotor,wolpert1995internal,rao2024sensory,clark2013whatever,ha2018world,lecun2022path,thorat2025predicting,nortmann2026predictive}.

\section{Acknowledgments}
The authors VB and TCK acknowledge funding by ERC StG grant 101039524 TIME. Compute resources for this project are in part funded by the Deutsche Forschungsgemeinschaft (DFG, German Research Foundation), project number 456666331.

We thank all members of the Kietzmann Lab at Osnabrück University, especially Philip Sulewski and Thomas Nortmann, for discussions that helped us get started on this endeavor and helped us improve the paper.

\bibliography{bib}

@article{elman1990finding,
  title={Finding structure in time},
  author={Elman, Jeffrey L},
  journal={Cognitive science},
  volume={14},
  number={2},
  pages={179--211},
  year={1990},
  publisher={Wiley Online Library}
}

@article{o2001sensorimotor,
  title={A sensorimotor account of vision and visual consciousness},
  author={O'regan, J Kevin and No{\"e}, Alva},
  journal={Behavioral and brain sciences},
  volume={24},
  number={5},
  pages={939--973},
  year={2001},
  publisher={Cambridge University Press}
}

@article{whittington2020tolman,
  title={The Tolman-Eichenbaum machine: unifying space and relational memory through generalization in the hippocampal formation},
  author={Whittington, James CR and Muller, Timothy H and Mark, Shirley and Chen, Guifen and Barry, Caswell and Burgess, Neil and Behrens, Timothy EJ},
  journal={Cell},
  volume={183},
  number={5},
  pages={1249--1263},
  year={2020},
  publisher={Elsevier}
}

@article{eperon2026action,
  title={Action information is integrated into entorhinal representations of conceptual space and is reflected in eye movements},
  author={Eperon, Alexander and Doeller, Christian F and Theves, Stephanie and Bottini, Roberto},
  journal={Plos Biology},
  volume={24},
  number={4},
  pages={e3003755},
  year={2026},
  publisher={Public Library of Science San Francisco, CA USA}
}

@article{scholte2025beyond,
  title={Beyond binding: from modular to natural vision},
  author={Scholte, H Steven and de Haan, Edward HF},
  journal={Trends in Cognitive Sciences},
  volume={29},
  number={6},
  pages={505--515},
  year={2025},
  publisher={Elsevier}
}

@article{ha2018world,
  title={World models},
  author={Ha, David and Schmidhuber, J{\"u}rgen},
  journal={arXiv preprint arXiv:1803.10122},
  volume={2},
  number={3},
  year={2018}
}

@inproceedings{petroni2019language,
  title={Language models as knowledge bases?},
  author={Petroni, Fabio and Rockt{\"a}schel, Tim and Riedel, Sebastian and Lewis, Patrick and Bakhtin, Anton and Wu, Yuxiang and Miller, Alexander},
  booktitle={Proceedings of the 2019 conference on empirical methods in natural language processing and the 9th international joint conference on natural language processing (EMNLP-IJCNLP)},
  pages={2463--2473},
  year={2019}
}

@article{brown2020language,
  title={Language models are few-shot learners},
  author={Brown, Tom and Mann, Benjamin and Ryder, Nick and Subbiah, Melanie and Kaplan, Jared D and Dhariwal, Prafulla and Neelakantan, Arvind and Shyam, Pranav and Sastry, Girish and Askell, Amanda and others},
  journal={Advances in neural information processing systems},
  volume={33},
  pages={1877--1901},
  year={2020}
}

@article{lecun2022path,
  title={A path towards autonomous machine intelligence version 0.9. 2, 2022-06-27},
  author={LeCun, Yann},
  journal={Open Review},
  volume={62},
  number={1},
  pages={1--62},
  year={2022}
}

@article{thorat2025predicting,
  title={Predicting upcoming visual features during eye movements yields scene representations aligned with human visual cortex},
  author={Thorat, Sushrut and Doerig, Adrien and Kroner, Alexander and Amme, Carmen and Kietzmann, Tim C},
  journal={arXiv preprint arXiv:2511.12715},
  year={2025}
}

@article{olsson2022context,
  title={In-context learning and induction heads},
  author={Olsson, Catherine and Elhage, Nelson and Nanda, Neel and Joseph, Nicholas and DasSarma, Nova and Henighan, Tom and Mann, Ben and Askell, Amanda and Bai, Yuntao and Chen, Anna and others},
  journal={arXiv preprint arXiv:2209.11895},
  year={2022}
}

@misc{cueva2018emergencegridlikerepresentationstraining,
      title={Emergence of grid-like representations by training recurrent neural networks to perform spatial localization}, 
      author={Christopher J. Cueva and Xue-Xin Wei},
      year={2018},
      eprint={1803.07770},
      archivePrefix={arXiv},
      primaryClass={q-bio.NC},
      url={https://arxiv.org/abs/1803.07770}, 
}

@article{banino2018vector,
  title={Vector-based navigation using grid-like representations in artificial agents},
  author={Banino, Andrea and Barry, Caswell and Uria, Benigno and Blundell, Charles and Lillicrap, Timothy and Mirowski, Piotr and Pritzel, Alexander and Chadwick, Martin J and Degris, Thomas and Modayil, Joseph and others},
  journal={Nature},
  volume={557},
  number={7705},
  pages={429--433},
  year={2018},
  publisher={Nature Publishing Group UK London}
}

@article{sorscher2023unified,
  title={A unified theory for the computational and mechanistic origins of grid cells},
  author={Sorscher, Ben and Mel, Gabriel C and Ocko, Samuel A and Giocomo, Lisa M and Ganguli, Surya},
  journal={Neuron},
  volume={111},
  number={1},
  pages={121--137},
  year={2023},
  publisher={Elsevier}
}

@article{treisman1980feature,
  title={A feature-integration theory of attention},
  author={Treisman, Anne M and Gelade, Garry},
  journal={Cognitive psychology},
  volume={12},
  number={1},
  pages={97--136},
  year={1980},
  publisher={Elsevier}
}

@article{kahneman1992reviewing,
  title={The reviewing of object files: Object-specific integration of information},
  author={Kahneman, Daniel and Treisman, Anne and Gibbs, Brian J},
  journal={Cognitive psychology},
  volume={24},
  number={2},
  pages={175--219},
  year={1992},
  publisher={Elsevier}
}

@article{hummel1992dynamic,
  title={Dynamic binding in a neural network for shape recognition.},
  author={Hummel, John E and Biederman, Irving},
  journal={Psychological review},
  volume={99},
  number={3},
  pages={480},
  year={1992},
  publisher={American Psychological Association}
}

@article{smolensky1990tensor,
  title={Tensor product variable binding and the representation of symbolic structures in connectionist systems},
  author={Smolensky, Paul},
  journal={Artificial intelligence},
  volume={46},
  number={1-2},
  pages={159--216},
  year={1990},
  publisher={Elsevier}
}

@article{plate1995holographic,
  title={Holographic reduced representations},
  author={Plate, Tony A},
  journal={IEEE Transactions on Neural networks},
  volume={6},
  number={3},
  pages={623--641},
  year={1995},
  publisher={IEEE}
}

@article{kanerva2009hyperdimensional,
  title={Hyperdimensional computing: An introduction to computing in distributed representation with high-dimensional random vectors},
  author={Kanerva, Pentti},
  journal={Cognitive computation},
  volume={1},
  number={2},
  pages={139--159},
  year={2009},
  publisher={Springer}
}

@article{kleyko2022survey,
  title={A survey on hyperdimensional computing aka vector symbolic architectures, part i: Models and data transformations},
  author={Kleyko, Denis and Rachkovskij, Dmitri A and Osipov, Evgeny and Rahimi, Abbas},
  journal={ACM Computing Surveys},
  volume={55},
  number={6},
  pages={1--40},
  year={2022},
  publisher={ACM New York, NY}
}

@incollection{mccloskey1989catastrophic,
  title={Catastrophic interference in connectionist networks: The sequential learning problem},
  author={McCloskey, Michael and Cohen, Neal J},
  booktitle={Psychology of learning and motivation},
  volume={24},
  pages={109--165},
  year={1989},
  publisher={Elsevier}
}

@article{mcclelland1995there,
  title={Why there are complementary learning systems in the hippocampus and neocortex: insights from the successes and failures of connectionist models of learning and memory.},
  author={McClelland, James L and McNaughton, Bruce L and O'Reilly, Randall C},
  journal={Psychological review},
  volume={102},
  number={3},
  pages={419},
  year={1995},
  publisher={American Psychological Association}
}

@article{rao2024sensory,
  title={A sensory--motor theory of the neocortex},
  author={Rao, Rajesh PN},
  journal={Nature neuroscience},
  volume={27},
  number={7},
  pages={1221--1235},
  year={2024},
  publisher={Nature Publishing Group US New York}
}

@article{clark2013whatever,
  title={Whatever next? Predictive brains, situated agents, and the future of cognitive science},
  author={Clark, Andy},
  journal={Behavioral and brain sciences},
  volume={36},
  number={3},
  pages={181--204},
  year={2013},
  publisher={Cambridge University Press}
}

@article{nortmann2026predictive,
  title={Predictive remapping and allocentric coding as consequences of energy efficiency in recurrent neural network models of active vision},
  author={Nortmann, Thomas and Sulewski, Philip and Kietzmann, Tim C},
  journal={Patterns},
  volume={7},
  number={1},
  year={2026},
  publisher={Elsevier}
}

@article{cho2014learning,
  title={Learning phrase representations using RNN encoder-decoder for statistical machine translation},
  author={Cho, Kyunghyun and Van Merri{\"e}nboer, Bart and Gulcehre, Caglar and Bahdanau, Dzmitry and Bougares, Fethi and Schwenk, Holger and Bengio, Yoshua},
  journal={arXiv preprint arXiv:1406.1078},
  year={2014}
}

@article{paszke2019pytorch,
  title={Pytorch: An imperative style, high-performance deep learning library},
  author={Paszke, Adam and Gross, Sam and Massa, Francisco and Lerer, Adam and Bradbury, James and Chanan, Gregory and Killeen, Trevor and Lin, Zeming and Gimelshein, Natalia and Antiga, Luca and others},
  journal={Advances in neural information processing systems},
  volume={32},
  year={2019}
}

@article{cortes1995support,
  title={Support-vector networks},
  author={Cortes, Corinna and Vapnik, Vladimir},
  journal={Machine learning},
  volume={20},
  number={3},
  pages={273--297},
  year={1995},
  publisher={Springer}
}

@article{pedregosa2011scikit,
  title={Scikit-learn: Machine learning in Python},
  author={Pedregosa, Fabian and Varoquaux, Ga{\"e}l and Gramfort, Alexandre and Michel, Vincent and Thirion, Bertrand and Grisel, Olivier and Blondel, Mathieu and Prettenhofer, Peter and Weiss, Ron and Dubourg, Vincent and others},
  journal={the Journal of machine Learning research},
  volume={12},
  pages={2825--2830},
  year={2011},
  publisher={JMLR. org}
}

@article{mccoy2018rnns,
  title={RNNs implicitly implement tensor product representations},
  author={McCoy, R Thomas and Linzen, Tal and Dunbar, Ewan and Smolensky, Paul},
  journal={arXiv preprint arXiv:1812.08718},
  year={2018}
}

@article{wolpert1995internal,
  title={An internal model for sensorimotor integration},
  author={Wolpert, Daniel M. and Ghahramani, Zoubin and Jordan, Michael I.},
  journal={Science},
  volume={269},
  number={5232},
  pages={1880--1882},
  year={1995},
  publisher={American Association for the Advancement of Science}
}

@article{wolpert2001motor,
  title={Motor prediction},
  author={Wolpert, Daniel M. and Flanagan, J. Randall},
  journal={Current Biology},
  volume={11},
  number={18},
  pages={R729--R732},
  year={2001},
  publisher={Elsevier}
}

@article{friston2011action,
  title={Action understanding and active inference},
  author={Friston, Karl and Mattout, J{\'e}r{\'e}mie and Kilner, James},
  journal={Biological Cybernetics},
  volume={104},
  pages={137--160},
  year={2011},
  publisher={Springer}
}

@article{greff2020binding,
  title={On the binding problem in artificial neural networks},
  author={Greff, Klaus and Van Steenkiste, Sjoerd and Schmidhuber, J{\"u}rgen},
  journal={arXiv preprint arXiv:2012.05208},
  year={2020}
}

@article{fodor1988connectionism,
  title={Connectionism and cognitive architecture: A critical analysis},
  author={Fodor, Jerry A and Pylyshyn, Zenon W},
  journal={Cognition},
  volume={28},
  number={1-2},
  pages={3--71},
  year={1988},
  publisher={Elsevier}
}

@article{griffiths2025whither,
  title={Whither symbols in the era of advanced neural networks?},
  author={Griffiths, Thomas L and Lake, Brenden M and McCoy, R Thomas and Pavlick, Ellie and Webb, Taylor W},
  journal={Trends in Cognitive Sciences},
  year={2025},
  publisher={Elsevier}
}

\end{document}